\documentclass[twocolumn,11pt]{template/servicenow}

\title{Are Online Skill and Memory Modules Always\\Worth Their Tokens?\\[0.5em]A Budget-Constrained Study of Web Agents}

\author[1,2]{Sina Hajimiri}
\author[1,2]{Masih Aminbeidokhti}
\author[2]{Jose Dolz}
\author[2]{Ismail Ben Ayed}
\author[1,3]{\\Issam H. Laradji}
\author[1,4,*]{Spandana Gella}
\author[1,*]{Nicolas Gontier}

\affiliation[1]{ServiceNow AI Research}
\affiliation[2]{ÉTS Montreal}
\affiliationbreak
\affiliation[3]{University of British Columbia}
\affiliation[4]{McGill University}
\contribution[*]{Equal advising}

\abstract{Online web agents often augment a base actor with memory, workflow, or skill modules.
These modules can improve performance, but they also consume test-time tokens, a cost rarely reported alongside the actor's inference cost.
We study online augmentation, where this overhead is paid on every task, and re-evaluate its benefits under a fixed total inference budget.
We compare AWM, ASI, and ReasoningBank with a token-matched vanilla baseline that uses the same budget for additional actor steps.
Across four WebArena domains and three models, Gemini~3~Flash, GPT-5.4-mini, and Qwen~3.6-27B, the vanilla baseline matches or surpasses all three augmentation methods in aggregate success rate while often using fewer total tokens.
We observe a similar trend on WorkArena-L1 with Qwen~3.6-27B, indicating that the effect extends to enterprise knowledge-work tasks.
Our results suggest that skills and workflow memory can be useful in specific domains, but their apparent gains often vanish against a budget-matched actor.
We further show that run-to-run variance materially affects outcomes and should be reported as a core evaluation criterion for online web agents.}

\website{\url{https://sinahmr.github.io/budget-constrained-web-agents}}

\usepackage{microtype}
\usepackage{inconsolata}
\usepackage{graphicx}

\usepackage{booktabs}
\usepackage{multirow}
\usepackage{array}
\usepackage{xcolor}
\usepackage{amsmath}
\usepackage{amssymb}
\usepackage{mathtools}
\usepackage{amsthm}
\usepackage{pifont}  
\usepackage{adjustbox}
\usepackage{graphicx}
\usepackage{tikz}
\usetikzlibrary{arrows.meta,positioning,calc,fit}
\usepackage{subcaption}

\usepackage{xurl}

\newcommand{\budgetactor}{\mbox{Vanilla-IB}}
\newcommand{\spara}[1]{\paragraph{#1}}

\begin{document}

\maketitle

\section{Introduction}

\begin{figure}[t]
    \centering
    \begin{subfigure}[t]{\linewidth}
     \centering
        \resizebox{\linewidth}{!}{
\definecolor{actorblue}{HTML}{0072B2}
\definecolor{moduleorange}{HTML}{D55E00}
\definecolor{promptred}{HTML}{CC6677}
\definecolor{plainteal}{HTML}{009E73}
\definecolor{softgray}{HTML}{F3F4F6}

\begin{tikzpicture}[
    font=\sffamily\footnotesize,
    >=Latex,
    box/.style={rounded corners=3pt, draw=black!35, very thick, fill=softgray, align=center},
    route/.style={rounded corners=3pt, draw=black!40, thick, fill=white, align=center},
    callout/.style={rounded corners=3pt, draw=black!35, fill=yellow!18, align=center, inner sep=4pt},
    lab/.style={align=center, inner sep=1pt},
    arr/.style={->, thick, draw=black!65},
]

\node[box, minimum width=10.4cm, minimum height=0.58cm] (budget) at (0,3.35)
{\textbf{Controlled total inference budget per task}};

\node[route, minimum width=4.85cm, minimum height=3.4cm] (aug) at (-2.75,1.25) {};
\node[route, minimum width=4.85cm, minimum height=3.4cm] (base) at (2.75,1.25) {};

\node[lab, font=\sffamily\bfseries\small] at (-2.75,2.65)
{Online augmentation};
\node[lab] at (-2.75,2.23)
{AWM / ASI / ReasoningBank};

\node[lab, font=\sffamily\bfseries\small] at (2.75,2.65)
{Budget-matched actor};
\node[lab] at (2.75,2.23)
{vanilla actor + more exploration};

\def\barw{3.95}
\def\barh{0.34}
\def\leftx{-4.73}
\def\rightx{0.78}
\def\bary{1.62}

\def\actorw{2.60}
\def\modulesw{1.35}

\draw[black!45, line width=0.45pt] (\leftx,\bary) rectangle ++(3.95,\barh);

\fill[actorblue!70] 
  (\leftx,\bary) rectangle ++(\actorw,\barh);

\fill[moduleorange!75] 
  ($(\leftx,\bary)+(\actorw,0)$) rectangle ++(\modulesw,\barh);

\node[white, font=\sffamily\scriptsize\bfseries] 
  at ($(\leftx,\bary)+(1.30,0.17)$) {actor};

\node[white, font=\sffamily\scriptsize\bfseries] 
  at ($(\leftx,\bary)+(3.275,0.17)$) {modules};

\node[lab, text width=4.30cm] at (-2.75,1.3)
{10 actor steps + module calls};

\draw[black!45, line width=0.45pt] (\rightx,\bary) rectangle ++(\barw,\barh);
\fill[actorblue!70] (\rightx,\bary) rectangle ++(\barw,\barh);
\node[white, font=\sffamily\scriptsize\bfseries] at ($(\rightx,\bary)+(1.98,0.17)$)
{actor};

\node[lab, text width=4.30cm] at (2.75,1.3)
{15 actor steps};

\node[callout, text width=3.95cm] at (-2.75,0.4)
{\textbf{Cost paid online:}\\ fewer tokens left for exploration};
\node[callout, text width=3.95cm] at (2.75,0.4)
{\textbf{Budget redirected:}\\ more interaction\\ with the environment};

\end{tikzpicture}}
    \end{subfigure}

    \vspace{0.2em}

    \begin{subfigure}[t]{\linewidth}
        \centering
        \includegraphics[width=0.95\linewidth]{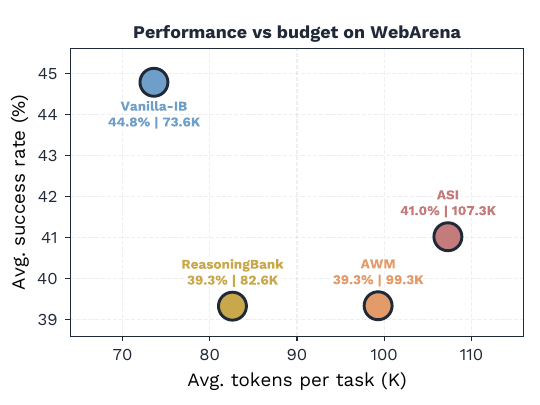}
    \end{subfigure}
\caption{
\textbf{Budget allocation under controlled test-time inference cost.}
(top) Online augmentation spends budget on workflow, skill, or memory modules, while a budget-matched vanilla actor (\budgetactor{}) spends it on additional observe-act steps.
(bottom) On WebArena with Gemini~3~Flash~\citep{gemini-team-2025-gemini3}, \budgetactor{} achieves the highest success rate with fewer tokens.
}
\label{fig:intro-budget}
\end{figure}

Web agents are evaluated on how often they solve tasks under a test-time budget.
A growing line of work augments a base actor with modules that induce reusable workflows, maintain memory, or distill skills from prior trajectories~\citep{wang-etal-2025-asi, wang-etal-2024-agent,  ouyang-etal-2025-reasoningbank, li-etal-2025-polyskill}.
Following Agent Skill Induction~\citep{wang-etal-2025-asi} and SkillWeaver~\citep{zheng-etal-2025-skillweaver}, we use the term \emph{skill} for executable code, typically a Python function, that wraps a reusable sequence of low-level actions.

These auxiliary modules are not free.
They consume tokens, add prompts and control logic, and in the online setting, couple performance to the order in which tasks are processed.
Their token cost is rarely reported next to the actor's, which makes it hard to tell whether observed gains are justified within a fixed budget.
This creates a simple allocation problem: tokens spent inducing, retrieving, verifying, or injecting reusable knowledge are tokens unavailable for direct task execution, such as observing the current page, reasoning over its state, or taking additional actions.

Figure~\ref{fig:intro-budget} illustrates this tradeoff.
Under the same total inference budget, an online augmented agent splits its budget between direct environment interaction and auxiliary memory, workflow, or skill machinery.
A budget-matched vanilla actor instead redirects that budget toward additional observe-act steps.
The bottom panel previews the main empirical pattern we observe on WebArena~\citep{zhou-etal-2024-webarena}: once total token usage is counted, a budget-matched vanilla agent achieves the highest average success rate while using fewer tokens than the online augmentation methods.

Prior work on augmentation methods largely assumed that base models lacked the capability to complete tasks end-to-end~\citep{wang-etal-2024-agent,wang-etal-2025-asi}; more recent frontier models, however, may no longer face that constraint and may require less external scaffolding.
We therefore ask a direct question: under a fixed inference budget, is it better to allocate tokens to online memory and skill modules, or can a vanilla actor achieve comparable performance by spending the same budget on additional interaction steps?
We focus on the \emph{online} setting, where agents update their behavior using trajectories collected during evaluation, rather than relying on offline skill libraries constructed before deployment.
In this setting, auxiliary costs are incurred repeatedly, the knowledge accumulated by memory or skill modules depends on the order in which tasks are encountered rather than the full task distribution, and task-level parallelism is constrained by the sequential dependence between tasks.

To study this tradeoff, we compare three online augmentation methods, Agent Workflow Memory (AWM; \citealp{wang-etal-2024-agent}), Agent Skill Induction (ASI; \citealp{wang-etal-2025-asi}), and ReasoningBank (\citealp{ouyang-etal-2025-reasoningbank}), against a budget-matched vanilla actor.
Unlike the augmented methods, which use the default 10-step actor horizon, this baseline allocates the same inference budget to a longer interaction horizon; we refer to it as \textbf{Vanilla}-\textbf{I}ncreased-\textbf{B}udget (\textbf{\budgetactor{}}).
We do not present \budgetactor{} as a new agent scaffold, but rather as a simple budget-aware control designed to test whether the overhead of online augmentation is justified.

We evaluate this comparison on WebArena~\citep{zhou-etal-2024-webarena} and WorkArena-L1~\citep{drouin-etal-2024-workarena}.
WebArena provides the broadest sweep in our study, spanning three models, four agent configurations, and four domains: Shopping, Reddit, Admin, and GitLab.
WorkArena-L1 tests the same budget-matching question in enterprise knowledge-work tasks under Qwen~3.6-27B.

Our main findings are as follows:
(1)~On WebArena, \budgetactor{} achieves the highest aggregate success rate across all three models we test, while mostly using fewer total tokens than AWM, ASI, and ReasoningBank.
(2)~On WorkArena-L1, \budgetactor{} remains competitive with the best-performing augmented methods, indicating that the budget-matching question is not specific to WebArena or consumer-style websites.
(3)~Beyond these aggregate results, we show that single-run success rate hides high task-level variance, and that reporting a method's performance from a single run is a weak basis for comparison.

These findings suggest three evaluation principles for online web agents. First, comparisons should report \emph{total token usage} across all modules, not only the actor.
Second, online augmentation methods should be compared against \emph{baselines within the same budget}.
Third, evaluations should report \emph{multi-run variance}, since agent evaluation introduces stochasticity that aggregate single-run metrics do not capture.

\section{Related work}

\spara{Web agent benchmarks.} Mind2Web~\citep{deng-etal-2024-mind2web} and WebArena~\citep{zhou-etal-2024-webarena} introduced large-scale benchmarks for realistic multi-step browser interaction. Subsequent work has broadened this benchmark landscape to cover visual grounding in VisualWebArena~\citep{koh-etal-2024-visualwebarena}, enterprise workflows in WorkArena~\citep{drouin-etal-2024-workarena} and WorkArena++~\citep{boisvert-etal-2024-workarenaplusplus}, live websites in Online-Mind2Web~\citep{xue-etal-2025-illusion}, and continuously generated open-ended tasks in WebArena-Infinity~\citep{webarena-infinity}. In this work, we focus on WebArena and WorkArena because they provide interactive environments in which agents must act, observe the consequences of their actions, and complete diverse tasks end-to-end. This makes them well suited for evaluating test-time compute mechanisms, where the cost of additional reasoning steps, tool calls, or memory access must be measured as part of the agent's deployment-time behavior.

\spara{Online workflow, memory, and skill induction.} A growing line of work augments base actors with modules that extract reusable knowledge from evaluation-time trajectories.
AWM~\citep{wang-etal-2024-agent} induces natural-language workflows from successful trajectories and retrieves them for later tasks.
ASI~\citep{wang-etal-2025-asi} synthesizes reusable Python skill functions called as higher-level operations for subsequent tasks.
ReasoningBank~\citep{ouyang-etal-2025-reasoningbank} distills reasoning strategies from both successes and failures into a growing memory bank queried at test time.
PolySkill~\citep{li-etal-2025-polyskill} separates a skill's abstract goal from its concrete realization for more general reuse, and PlugMem~\citep{yang-etal-2026-plugmem} proposes a task-agnostic memory plugin attachable to any LLM agent.
ACE~\citep{zhang-etal-2025-ace} treats the agent's context itself as an evolving playbook that accumulates and refines strategies through a generate-reflect-curate loop.
ERL~\citep{allard-etal-2026-erl} reflects on task trajectories to extract transferable heuristics, which are retrieved at test time to guide execution on subsequent tasks.
All these methods share a common premise: part of the test-time budget is allocated to producing and retrieving reusable knowledge, a cost that is rarely reported alongside the actor's.
Our work asks whether that allocation is always the right choice.

\spara{Offline reusable knowledge.} A complementary line of work builds reusable knowledge before deployment begins, through an exploration or synthesis phase that is separate from the evaluation run.
SkillWeaver~\citep{zheng-etal-2025-skillweaver} discovers and synthesizes Python API skills in an offline phase before deployment.
WALT~\citep{prabhu-etal-2025-walt} learns stable tool abstractions reverse-engineered from website structure prior to evaluation, and
WebXSkill~\citep{wang-etal-2026-webxskill} mines parameterized action programs from synthetic agent trajectories.
JEF-Hinter~\citep{nekoei-etal-2025-jefhinter} distills offline trajectories into compact, context-aware hints, including from both successful and failed traces, and retrieves relevant hints for the agent's current state at inference time.
Offline approaches amortize discovery or distillation cost over many subsequent uses, which is useful when an agent is deployed repeatedly on the same stable environment.
This tradeoff breaks down when the environment is visited infrequently, when its structure changes between discovery and deployment, or when the agent encounters a new domain.
These limitations motivate the online setting that is the focus of our work.

\spara{LLM capability and the value of scaffolding.} Frontier language models have rapidly improved in capabilities central to web agents: multi-step reasoning, instruction following, and adaptive decision-making under uncertainty.
Recent models such as GPT-5.4~\citep{openai-2026-gpt54}, Claude Sonnet~4.6~\citep{anthropic-2026-sonnet46}, and Gemini~3~\citep{gemini-team-2025-gemini3} can solve complex long-horizon tasks with relatively lightweight scaffolding, often matching or outperforming more heavily engineered agent architectures~\citep{bechard-etal-2026-terminal}.
This rapid progress has consequences for augmentation methods: gains that were genuine and substantial over weaker models may shrink or disappear as the base actor becomes more capable on its own.

\spara{Alternative paradigms and evaluation robustness.}
Most web agents use a browser, accessibility-tree, and fixed action vocabulary.
A line of work explores structurally different paradigms: API-based agents substantially outperform browser agents on WebArena, with hybrid agents reaching even higher performance~\citep{song-etal-2025-beyond}, and terminal-based agents operating through code and programmatic interfaces can match or outperform GUI-based agents on enterprise automation tasks~\citep{bechard-etal-2026-terminal}.
These approaches avoid some browser-skill fragilities, but depend on stable, documented APIs that may not exist.
Within the browser-based setting, evaluation reliability is itself a concern: timeouts, page-load failures, and non-deterministic UI behavior can change outcomes independently of agent decisions. Introducing realistic network failures into WebArena, WebVoyager~\citep{he-etal-2024-webvoyager}, and REAL~\citep{garg2025realbenchmarkingautonomousagents} has been shown to cause significant drops in success rates~\citep{kara-etal-2025-warex}.
These issues reinforce the importance of multi-run reporting as a first-class evaluation criterion.

\section{Experimental settings}
\label{sec:setup}

\subsection{Tasks and domains}

We evaluate on WebArena~\citep{zhou-etal-2024-webarena}, a benchmark of realistic multi-step web tasks across isolated website environments.
We report results on four domains.
\textbf{Shopping}, consisting of 187 tasks, is an e-commerce site with product search, comparison, and checkout tasks.
\textbf{Reddit}, with 106 tasks, is a social discussion platform with navigation, posting, and search tasks.
\textbf{Admin}, which contains 182 tasks, is the administrator panel of the WebArena shopping site, covering product management, order processing, and account administration.
\textbf{GitLab}, with 180 tasks, is a collaborative software-development platform with repository browsing, issue tracking, merge-request, and project management tasks.
These domains differ in how much their tasks share navigational patterns, which is relevant to whether workflow or skill reuse can pay off.
We do not evaluate on WebArena's Map domain due to website reliability issues, following prior work~\citep{ouyang-etal-2025-reasoningbank,miyai-etal-2025-webchorearena}.

The setting is sequential and online: agents attempt tasks in order, and use modules with skill, memory, or workflow components to extract knowledge from earlier trajectories and apply it to later ones.

We also evaluate on the Level-1 collection of WorkArena~\citep{drouin-etal-2024-workarena} tasks focused on common enterprise knowledge-work interactions in the ServiceNow platform.
This benchmark consists of 33 task types spanning ServiceNow operations such as dashboards, forms, knowledge-base search, list filtering, list sorting, menu navigation, and service-catalog requests.
For WorkArena-L1, we conduct the same online comparison across all task types, using three seeds per task type for each method.
This benchmark replaces consumer-style and community websites with enterprise software workflows, while preserving the browser-based interaction setting used by our agents.

\subsection{Studied systems}

We compare four agent scaffolds in our experiments. Three methods augment the actor with auxiliary modules while keeping the actor's interaction horizon to at most 10 steps. The fourth, \budgetactor{}, is a vanilla actor without auxiliary modules but with an extended interaction budget.

\noindent\textbf{AWM}~\citep{wang-etal-2024-agent}: online workflow memory, in which an LLM-based auxiliary module induces and retrieves workflows from prior trajectories alongside the actor.

\noindent\textbf{ASI}~\citep{wang-etal-2025-asi}: online programmatic skill induction, in which an LLM-based auxiliary module synthesizes, verifies, and reuses Python skill functions.

\noindent\textbf{ReasoningBank}~\citep{ouyang-etal-2025-reasoningbank}: online reasoning memory, in which an LLM-based auxiliary module accumulates and retrieves reasoning strategies from prior successes and failures (denoted RBank in tables).

\noindent\textbf{\budgetactor{}}: a vanilla actor with its interaction horizon extended to 15 steps and with rule-based pruning of the accessibility-tree.
The pruning removes text duplicated between a parent node and its direct children, a common pattern in accessibility-trees that inflates context length without adding information.
This pruning involves no LLM calls (see Appendix~\ref{app:pruning}).
We do not present these modifications as a contribution; they are included only to give the vanilla actor more interaction steps under a comparable overall budget.

The 15-step horizon for \budgetactor{} typically does not exceed the total token budget of the augmented methods, because their auxiliary modules consume a substantial fraction of the overall tokens.
Exact budget matching is difficult in exploratory web agent benchmarks such as WebArena and WorkArena.
Token usage varies across tasks because agents choose actions dynamically and task complexity differs, so there is no fixed per-task cost to target.
Moreover, for augmented methods, a single fixed budget cannot be cleanly divided between the actor and its auxiliary modules.
The augmented methods in our comparison also use different aggregate budgets, so there is no single token value against which to match \budgetactor{}. The most practical control lever is therefore the actor's maximum step count.
Setting this horizon to 15 is an approximation rather than an exact budget match.
In many tasks, \budgetactor{} could take additional steps while still remaining within the token budget of one or more augmented baselines.
In a small number of tasks, the reverse is true: \budgetactor{} spends slightly more tokens than one or more augmented methods.
We acknowledge this as a limitation of step-count-based budget control, although such cases are uncommon in our experiments.

\begin{table*}[t]
\centering
\resizebox{\textwidth}{!}{
\begin{tabular}{lcccccccccc}
\toprule
 & \multicolumn{2}{c}{\textbf{Shopping} (187)} & \multicolumn{2}{c}{\textbf{Reddit} (106)} & \multicolumn{2}{c}{\textbf{Admin} (182)} & \multicolumn{2}{c}{\textbf{GitLab} (180)} & \multicolumn{2}{c}{\textbf{Avg.}} \\
\cmidrule(lr){2-3}\cmidrule(lr){4-5}\cmidrule(lr){6-7}\cmidrule(lr){8-9}\cmidrule(lr){10-11}
\textbf{Method} & \textbf{SR $\uparrow$} & \textbf{Tok $\downarrow$} & \textbf{SR $\uparrow$} & \textbf{Tok $\downarrow$} & \textbf{SR $\uparrow$} & \textbf{Tok $\downarrow$} & \textbf{SR $\uparrow$} & \textbf{Tok $\downarrow$} & \textbf{SR $\uparrow$} & \textbf{Tok $\downarrow$} \\
\midrule
\multicolumn{11}{c}{\textbf{Gemini~3~Flash}} \\
\midrule
\budgetactor{} & \textbf{47.77}{\scriptsize $\pm$1.9} & \textbf{45.7}{\scriptsize $\pm$0.8}  & \textbf{47.48}{\scriptsize $\pm$0.5} & \textbf{71.7}{\scriptsize $\pm$2.2}  & \textbf{55.68}{\scriptsize $\pm$1.6} & \textbf{99.0}{\scriptsize $\pm$4.0}  & \textbf{29.07}{\scriptsize $\pm$0.6} & 78.0{\scriptsize $\pm$1.8}  & \textbf{44.78} & \textbf{73.6}  \\
AWM       & 41.18{\scriptsize $\pm$2.1} & 70.8{\scriptsize $\pm$3.0}  & \textbf{47.48}{\scriptsize $\pm$1.4} & 87.2{\scriptsize $\pm$2.0}  & 47.43{\scriptsize $\pm$0.6} & 142.7{\scriptsize $\pm$5.4}  & 24.44{\scriptsize $\pm$1.0} & 92.3{\scriptsize $\pm$1.1}  & 39.34 & 99.3 \\
ASI       & 44.74{\scriptsize $\pm$1.2} & 82.8{\scriptsize $\pm$0.3}  & 44.97{\scriptsize $\pm$2.0} & 94.6{\scriptsize $\pm$4.8}  & 52.75{\scriptsize $\pm$1.1} & 139.4{\scriptsize $\pm$11.2} & 22.96{\scriptsize $\pm$2.0} & 107.7{\scriptsize $\pm$5.8} & 41.02 & 107.3 \\
RBank     & 45.45{\scriptsize $\pm$1.4} & 54.6{\scriptsize $\pm$1.3}  & 39.93{\scriptsize $\pm$1.4} & 76.8{\scriptsize $\pm$2.8}  & 48.90{\scriptsize $\pm$1.9} & 124.6{\scriptsize $\pm$3.3}  & 22.96{\scriptsize $\pm$0.9} & \textbf{72.7}{\scriptsize $\pm$1.0}  & 39.33 & 82.6 \\
\midrule
\multicolumn{11}{c}{\textbf{GPT-5.4-mini}} \\
\midrule
\budgetactor{} & \textbf{38.50}{\scriptsize $\pm$0.9} & \textbf{54.3}{\scriptsize $\pm$0.8}  & \textbf{34.59}{\scriptsize $\pm$4.8} & 96.2{\scriptsize $\pm$6.6}  & \textbf{35.89}{\scriptsize $\pm$1.3} & 124.1{\scriptsize $\pm$8.0}  & \textbf{22.22}{\scriptsize $\pm$2.2} & 89.8{\scriptsize $\pm$4.4}  & \textbf{32.67} & 90.2 \\
AWM       & 29.95{\scriptsize $\pm$1.9} & 62.0{\scriptsize $\pm$2.0}  & 29.56{\scriptsize $\pm$1.1} & 80.4{\scriptsize $\pm$12.6} & 32.24{\scriptsize $\pm$3.2} & \textbf{113.4}{\scriptsize $\pm$16.8} & 17.22{\scriptsize $\pm$1.1} & 95.7{\scriptsize $\pm$2.7}  & 27.02 & 88.5 \\
ASI       & 33.15{\scriptsize $\pm$0.9} & 73.4{\scriptsize $\pm$9.0}  & 28.93{\scriptsize $\pm$2.2} & 87.2{\scriptsize $\pm$4.9}  & 32.96{\scriptsize $\pm$3.8} & 138.5{\scriptsize $\pm$10.9} & 20.74{\scriptsize $\pm$0.6} & 95.7{\scriptsize $\pm$1.9}  & 29.00 & 99.8 \\
RBank     & 26.38{\scriptsize $\pm$3.9} & 57.9{\scriptsize $\pm$1.6}  & 22.01{\scriptsize $\pm$2.9} & \textbf{69.2}{\scriptsize $\pm$2.9}  & 34.25{\scriptsize $\pm$1.8} & 123.1{\scriptsize $\pm$2.3}  & 14.45{\scriptsize $\pm$2.9} & \textbf{69.3}{\scriptsize $\pm$2.2}  & 24.58 & \textbf{81.0} \\
\midrule
\multicolumn{11}{c}{\textbf{Qwen~3.6-27B}} \\
\midrule
\budgetactor{} & \textbf{45.99}{\scriptsize $\pm$1.4} & \textbf{58.8}{\scriptsize $\pm$1.5}  & \textbf{44.34}{\scriptsize $\pm$2.5} & \textbf{89.7}{\scriptsize $\pm$4.3}  & \textbf{50.73}{\scriptsize $\pm$1.4} & \textbf{130.3}{\scriptsize $\pm$9.5}  & \textbf{28.15}{\scriptsize $\pm$1.2} & 101.8{\scriptsize $\pm$6.5}  & \textbf{42.14} & \textbf{95.5} \\
AWM       & 41.35{\scriptsize $\pm$1.9} & 71.8{\scriptsize $\pm$4.5}  & 43.08{\scriptsize $\pm$3.0} & 103.6{\scriptsize $\pm$13.8} & 46.15{\scriptsize $\pm$3.1} & 178.5{\scriptsize $\pm$12.5} & 23.33{\scriptsize $\pm$1.1} & 102.2{\scriptsize $\pm$4.8} & 38.01 & 115.0 \\
ASI       & 43.67{\scriptsize $\pm$3.5} & 78.8{\scriptsize $\pm$5.5}  & 43.08{\scriptsize $\pm$1.4} & 120.5{\scriptsize $\pm$4.7}  & 49.08{\scriptsize $\pm$2.3} & 182.2{\scriptsize $\pm$4.2}  & 25.74{\scriptsize $\pm$1.4} & 120.6{\scriptsize $\pm$3.6} & 40.15 & 125.8 \\
RBank     & 39.75{\scriptsize $\pm$0.8} & 65.4{\scriptsize $\pm$0.2}  & 41.19{\scriptsize $\pm$1.4} & 99.1{\scriptsize $\pm$5.4}  & 47.62{\scriptsize $\pm$1.7} & 156.9{\scriptsize $\pm$8.6}  & 20.93{\scriptsize $\pm$1.7} & \textbf{86.0}{\scriptsize $\pm$5.7}  & 37.00 & 101.9 \\
\bottomrule
\end{tabular}
}
\caption{\textbf{Per-domain success rate (\%) and total token usage per task (K) on WebArena.}
Domain values are means over three independent runs (mean~$\pm$~{\scriptsize std}); per-run details are in Appendix~\ref{app:variance}. Avg.\ is the task-weighted mean over tasks of all four domains.
}
\label{tab:main}
\end{table*}

\subsection{Models and budget accounting}

We evaluate under three models, \textsc{Gemini~3~Flash}~\citep{gemini-team-2025-gemini3}, \textsc{GPT-5.4-mini}~\citep{openai-2026-gpt54mini}, and \textsc{Qwen~3.6-27B}~\citep{qwen-team-2025-qwen36}, each across three independent runs on each domain.
On WorkArena-L1, we report the results under \textsc{Qwen~3.6-27B}.
For every configuration, we report task \textbf{success rate (SR)} and \textbf{total token usage}, which aggregates all LLM calls made by the system: actor and any auxiliary module calls (\textit{i.e.}, workflow induction, skill synthesis, memory construction, retrieval, and verification).
Token usage is averaged per task in thousands of tokens~(K).
For both WebArena and WorkArena-L1, metrics are reported as means and standard deviations over three runs.
Prior work typically reports metrics on one run, and often does not report token usage; we treat this accounting as a first-class evaluation criterion.

\section{Main results}
\label{sec:main}

\spara{Performance and token usage.} Table~\ref{tab:main} summarizes WebArena success rate and total token usage across domains, methods, and models.
Across WebArena domains, \budgetactor{} achieves the best average success rate for all three base models and, for Gemini~3~Flash and Qwen~3.6-27B, is also the most token-efficient configuration.
The augmented methods therefore do not simply trade extra tokens for higher accuracy; their workflow, skill, or memory modules often increase cost without producing proportional gains in task success.
This complicates the interpretation of scaffolded-agent gains: auxiliary reasoning can make a system appear stronger by giving it more computation, but much of that advantage disappears when compared with a vanilla actor that spends a similar budget on additional environment interaction.
The effect is strongest for GPT-5.4-mini, suggesting that scaffold quality is itself model-dependent: weaker models can produce noisy workflows, skills, or reasoning memories that are difficult for the actor to ignore.
For stronger models, the augmented methods are less prone to bad insights but still do not overtake \budgetactor{}, consistent with modern actors already internalizing many of the planning and navigation abilities that earlier scaffolds attempted to externalize.

\begin{figure}[bt]
\centering
\includegraphics[width=\columnwidth]{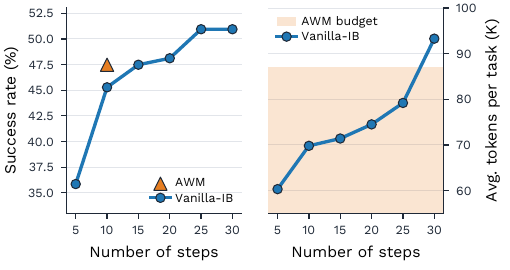}
\caption{\textbf{\budgetactor{} horizon sweep on Reddit under Gemini~3~Flash.} Success rate (left) and total tokens per task (right) are reported as a function of interaction steps. The available budget is the token usage of the best-performing augmentation baseline, in this case AWM, shaded in orange (right). At the 10-step mark, \mbox{\budgetactor{}} SR is below AWM, which in isolation might seem to justify augmentation. However, the right panel shows AWM already spends about 25\% more tokens at that point. Extending the \budgetactor{} horizon up to 25 steps closes this gap: \budgetactor{} surpasses AWM in SR while still using fewer tokens than AWM.
}
\label{fig:sweep}
\end{figure}

\spara{Reddit step-sweep.}
The closest case for augmentation is Reddit with Gemini~3~Flash, where AWM and \budgetactor{} have identical success rates.
We probe this case by extending the \budgetactor{} actor horizon further (Figure~\ref{fig:sweep}).
As the actor horizon increases, \budgetactor{} steadily improves, matching AWM from 15 steps onward and surpassing it at larger horizons while still using fewer tokens up until 25 steps.
This comparison also clarifies the role of step count: the augmented methods already exceed the 15-step \budgetactor{} token budget at their standard 10-step horizon, so giving them the same longer horizon would increase rather than resolve the cost imbalance.
The apparent benefit of workflow memory on Reddit is therefore at least partly explained by budget asymmetry rather than by a robust advantage of the auxiliary module itself.

\begin{table}[t]
\centering
\small
\begin{adjustbox}{width=0.4\textwidth}{
\begin{tabular}{lcc}
\toprule
\textbf{Method} & \textbf{SR $\uparrow$} & \textbf{Tok $\downarrow$} \\
\midrule
\budgetactor{} & \textbf{55.56}{\scriptsize $\pm$2.9} & 109.4{\scriptsize $\pm$14.8} \\
AWM       & 53.53{\scriptsize $\pm$3.8} & \textbf{102.8}{\scriptsize $\pm$2.6} \\
ASI       & 48.49{\scriptsize $\pm$4.3} & 113.8{\scriptsize $\pm$10.2} \\
RBank     & \textbf{55.56}{\scriptsize $\pm$2.8} & 120.3{\scriptsize $\pm$5.7} \\
\bottomrule
\end{tabular}
}\end{adjustbox}
\caption{\textbf{Success rate (\%) and total token usage per task (K) on WorkArena-L1} (33 task types) for Qwen~3.6-27B. Values are means over three independent seeds per task type.}
\label{tab:main:workarena}
\end{table}

\spara{WorkArena-L1.} The enterprise setting is more balanced but supports the same basic conclusion.
Table~\ref{tab:main:workarena} shows that, under Qwen~3.6-27B, \mbox{\budgetactor{}} is effectively tied with ReasoningBank for the best success rate, outperforms AWM and ASI, and uses fewer tokens than ASI and ReasoningBank.
Thus, \budgetactor{} remains on the success--cost frontier rather than being dominated by an augmented method.

\spara{Where does the token budget go?}
To identify the source of the token overhead, we decompose total usage into actor tokens from environment interaction and module tokens from workflow induction, skill synthesis, retrieval, and verification, in Table~\ref{tab:token-breakdown}.
The overhead has two sources.
First, augmented methods pay explicit module costs.
Second, and less obviously, augmentation methods also inflate the actor's prompt by injecting retrieved workflows, memories, or available skill functions at each step.
This implicit cost cannot be removed by making module calls cheaper: as long as retrieved knowledge is inserted into the actor context, the per-step actor cost rises with the size of that content.
Augmentation therefore imposes a double cost, combining auxiliary inference with larger actor prompts.

\begin{table*}[tb]
\centering
\resizebox{\textwidth}{!}{
\begin{tabular}{l cc cc cc cc cc cc cc cc}
\toprule
 & \multicolumn{4}{c}{\textbf{Shopping}} & \multicolumn{4}{c}{\textbf{Reddit}} & \multicolumn{4}{c}{\textbf{Admin}} & \multicolumn{4}{c}{\textbf{GitLab}} \\
\cmidrule(lr){2-5}\cmidrule(lr){6-9}\cmidrule(lr){10-13}\cmidrule(lr){14-17}
 & \multicolumn{2}{c}{Actor} & \multicolumn{2}{c}{Modules} & \multicolumn{2}{c}{Actor} & \multicolumn{2}{c}{Modules} & \multicolumn{2}{c}{Actor} & \multicolumn{2}{c}{Modules} & \multicolumn{2}{c}{Actor} & \multicolumn{2}{c}{Modules} \\
\cmidrule(lr){2-3}\cmidrule(lr){4-5}\cmidrule(lr){6-7}\cmidrule(lr){8-9}\cmidrule(lr){10-11}\cmidrule(lr){12-13}\cmidrule(lr){14-15}\cmidrule(lr){16-17}
\textbf{Method} & P. & C. & P. & C. & P. & C. & P. & C. & P. & C. & P. & C. & P. & C. & P. & C. \\
\midrule
\budgetactor{} & 45.2 & 0.5 & 0 & 0 & 71.3 & 0.4 & 0 & 0 & 98.4 & 0.6 & 0 & 0 & 77.3 & 0.7 & 0 & 0 \\
AWM       & 64.1 & 0.4 & 6.1 & 0.2 & 82.9 & 0.4 & 3.7 & 0.2 & 135.0 & 0.5 & 6.9 & 0.3 & 86.1 & 0.6 & 5.4 & 0.2 \\
ASI       & 61.8 & 0.5 & 20.0 & 0.5 & 84.6 & 0.4 &  9.3 & 0.3 & 118.1 & 0.5 & 20.2 & 0.6 & 85.2 & 0.5 & 21.4 & 0.5 \\
RBank         & 36.6 & 1.1 & 16.5 & 0.4 & 53.3 & 1.0 & 22.1 & 0.4 & 96.9 & 1.2 & 26.1 & 0.4 & 51.3 & 1.8 & 19.2 & 0.4 \\
\bottomrule
\end{tabular}
}
\caption{\textbf{Token breakdown by component under Gemini~3~Flash on WebArena}, per task in thousands (K). Actor: tokens from LLM calls made during interaction steps. Modules: tokens from auxiliary components (workflow induction, skill synthesis, retrieval, verification). Both are further split into prompt (\textit{P.}) and completion (\textit{C.}) tokens. Values are means over three runs. \budgetactor{} has no module calls by construction.}
\label{tab:token-breakdown}
\end{table*}

\spara{Headroom for \budgetactor{}.} We set the \budgetactor{} horizon to 15 steps as a clean, domain-agnostic choice, but in most cases it is a conservative setting.
As the Reddit sweep shows (Figure~\ref{fig:sweep}), token usage at 15 steps is still well below AWM's budget, and success rate continues to rise before saturating around 25 steps.
The unused headroom is even larger on Shopping, where the token gap to augmentation methods is wider.
Together, these results suggest that comparisons should be made along the axis of total token cost rather than nominal step count. AWM evaluated at 10 steps already spends more tokens than \budgetactor{} at 15 steps, so extending its actor horizon would only widen this cost gap rather than close it. For any step budget at which augmented methods are evaluated, a budget-matched vanilla baseline can be obtained by increasing the \budgetactor{} horizon accordingly; the same token-efficiency imbalance therefore persists across operating points.
We confirm this directly by re-running all methods at a higher budget and observe that the accuracy--cost ordering is preserved (Appendix~\ref{app:step-budget}).

\section{Audit beyond aggregate success rate}
\label{sec:audit}

\subsection{Run-to-run variance}
\label{sec:variance}

\spara{Multi-run reporting.} A single-run aggregate success rate hides properties that matter for online web agents.
Even at a fixed temperature, LLM outputs vary across runs due to hardware-level non-determinism in distributed inference.
A given task may succeed in one run and fail in another for reasons unrelated to agent capability: a different token sample produces a different action sequence, which may or may not navigate the same interface in the same way.
Another source of variance can be the web environment itself: BrowserGym~\citep{chezelles-etal-2025-browsergym} and Playwright~\citep{playwright} can encounter timeouts, slow page loads, or intermittent server errors, and prior work shows that injected web failures can sharply reduce success rates on web tasks~\citep{kara-etal-2025-waber, kara-etal-2025-warex}.
As a result, two runs of the same agent on the same task set can produce nearly identical aggregate success rates while disagreeing on a meaningful fraction of individual tasks.
This aggregate stability is reassuring for reproducibility but hides large task-level noise that single-run evaluations do not expose.

Multi-run reporting helps absorb both sources of noise and provides a more reliable picture of what the agent can actually do.
A single reported success rate is an incomplete statistic: it conflates genuine agent capability with the particular set of environment states encountered in that run, and this conflation is especially dangerous when comparing methods that differ in the number or structure of their LLM calls, since some methods have more opportunities to encounter a bad environment state. The effect is concrete even for the step-matched comparison that motivates augmentation: at a matched interaction budget, a plain vanilla actor can beat a given augmented method in one run and lose to it in another, so a single-run comparison can support opposite headlines. We therefore perform our experiments over three runs, and the detailed per-run results are provided in Appendix~\ref{app:variance}.

\spara{Any-of-3 and All-of-3.} To expose the task-level instability, we compute two complementary bounds:
\emph{Any-of-3}, the fraction of tasks that succeed in at least one of the three runs (an OR across runs); and
\emph{All-of-3}, the fraction that succeed in all three runs (an AND across runs).
Table~\ref{tab:any-all-3} reports both as illustrative bounds on the latent capability that single-run averages hide, and shows that the gaps are relatively large for all the methods.
Single-run success rate is therefore an incomplete estimate of task-level capability, especially for methods whose learned state changes across the evaluation sequence.

\begin{table*}[tb]
\centering
\resizebox{\textwidth}{!}{
\begin{tabular}{l cccc cccc cccc}
\toprule
 & \multicolumn{4}{c}{\textbf{Gemini~3~Flash}} & \multicolumn{4}{c}{\textbf{GPT-5.4-mini}} & \multicolumn{4}{c}{\textbf{Qwen~3.6-27B}} \\
\cmidrule(lr){2-5}\cmidrule(lr){6-9}\cmidrule(lr){10-13}
\textbf{Method} & \textbf{Any} & $\boldsymbol{\Delta_\uparrow}$ & \textbf{All} & $\boldsymbol{\Delta_\downarrow}$ & \textbf{Any} & $\boldsymbol{\Delta_\uparrow}$ & \textbf{All} & $\boldsymbol{\Delta_\downarrow}$ & \textbf{Any} & $\boldsymbol{\Delta_\uparrow}$ & \textbf{All} & $\boldsymbol{\Delta_\downarrow}$ \\
\midrule
\budgetactor{} & \textbf{54.01} & +6.2 & \textbf{42.25} & $-$5.5 & \textbf{46.52} & +8.0 & \textbf{27.81} & $-$10.7 & \textbf{54.01} & +8.0 & \textbf{38.50} & $-$7.5 \\
AWM       & 47.06 & +5.9 & 35.83 & $-$5.4 & 37.43 & +7.5 & 21.39 & $-$8.6 & 49.73 & +8.4 & 33.16 & $-$8.2 \\
ASI       & 51.34 & +6.6 & 37.43 & $-$7.3 & 40.64 & +7.5 & 26.74 & $-$6.4 & 52.41 & +8.7 & 36.36 & $-$7.3 \\
RBank     & 53.48 & +8.0 & 38.50 & $-$7.0 & 35.29 & +8.9 & 17.64 & $-$8.7 & 47.06 & +7.3 & 34.22 & $-$5.5 \\
\bottomrule
\end{tabular}
}
\caption{\textbf{Any-of-3 and All-of-3 success rates (\%) on the Shopping domain of WebArena} across three models. Any-of-3 is the fraction of tasks succeeding in at least one run; All-of-3 is the fraction succeeding in all three runs. $\Delta_\uparrow$ and $\Delta_\downarrow$ are the gaps from the three-run mean (Table~\ref{tab:main}).}
\label{tab:any-all-3}
\end{table*}

\subsection{Programmatic skill fragility}
\label{sec:fragility}

A structural limit of programmatic skill approaches such as ASI is that a synthesized skill is tied to the accessibility-tree state in which it was created.
These skills often wrap a sequence of low-level actions and refer to UI elements by BrowserGym accessibility-tree IDs (BIDs), which are assigned dynamically.
BIDs are not stable across interactions. After the first action in a skill, the page may reload or update, and later BIDs may then point to different elements or to nothing at all. The function then either fails silently, throws an error that disrupts the agent's trajectory, or interacts with the wrong element.

A vanilla agent that queries the LLM at each step adapts to state changes naturally, since it observes the current accessibility-tree before choosing each action.
A function-based approach lacks this adaptability unless it inserts an extra LLM call inside the function at each sub-step, in which case its efficiency justification largely disappears.
Appendix~\ref{app:failure-modes} provides explanations of this and related failure modes across all three augmented methods, including concrete task-level examples where augmented agents fail and \budgetactor{} succeeds.

\subsection{Effect of accessibility-tree pruning}
\label{sec:pruning-ablation}

\budgetactor{} combines two changes relative to a plain 10-step agent: an extended 15-step actor horizon and rule-based accessibility-tree pruning (Appendix~\ref{app:pruning}).
Table~\ref{tab:pruning-ablation} isolates their individual contributions.

\begin{table*}[bt]
\centering
\begin{adjustbox}{width=\textwidth}{
\begin{tabular}{lcccccccc}
\toprule
 & \multicolumn{2}{c}{\textbf{Shopping}} & \multicolumn{2}{c}{\textbf{Reddit}} & \multicolumn{2}{c}{\textbf{Admin}} & \multicolumn{2}{c}{\textbf{GitLab}} \\
\cmidrule(lr){2-3}\cmidrule(lr){4-5}\cmidrule(lr){6-7}\cmidrule(lr){8-9}
\textbf{Configuration} & \textbf{SR $\uparrow$} & \textbf{Tok $\downarrow$} & \textbf{SR $\uparrow$} & \textbf{Tok $\downarrow$} & \textbf{SR $\uparrow$} & \textbf{Tok $\downarrow$} & \textbf{SR $\uparrow$} & \textbf{Tok $\downarrow$} \\
\midrule
10 steps         & 45.81{\scriptsize $\pm$1.5} & 48.0{\scriptsize $\pm$1.1}  & 45.28{\scriptsize $\pm$1.0} & 75.2{\scriptsize $\pm$6.4}  & 48.72{\scriptsize $\pm$0.3} & 103.5{\scriptsize $\pm$7.7}  & 27.22{\scriptsize $\pm$0.6} & 65.6{\scriptsize $\pm$0.6} \\
10 steps, pruned & 46.35{\scriptsize $\pm$2.2} & \textbf{41.7}{\scriptsize $\pm$0.5}  & 47.17{\scriptsize $\pm$0.9} & \textbf{70.8}{\scriptsize $\pm$1.6}  & 51.65{\scriptsize $\pm$1.0} & \textbf{85.5}{\scriptsize $\pm$3.9}  & 27.04{\scriptsize $\pm$1.4} & \textbf{60.9}{\scriptsize $\pm$0.8}  \\
15 steps         & 47.24{\scriptsize $\pm$2.8} & 54.0{\scriptsize $\pm$1.1}  & \textbf{47.48}{\scriptsize $\pm$2.2} & 77.7{\scriptsize $\pm$6.7}  & \textbf{57.14}{\scriptsize $\pm$0.0} & 119.8{\scriptsize $\pm$16.1} & \textbf{30.56}{\scriptsize $\pm$0.0} & 78.0{\scriptsize $\pm$2.1} \\
15 steps, pruned & \textbf{47.77}{\scriptsize $\pm$1.9} & 45.7{\scriptsize $\pm$0.8}  & \textbf{47.48}{\scriptsize $\pm$0.5} & 71.7{\scriptsize $\pm$2.2}  & 55.68{\scriptsize $\pm$1.6} & 99.0{\scriptsize $\pm$4.0}  & 29.07{\scriptsize $\pm$0.6} & 78.0{\scriptsize $\pm$1.8}  \\
\bottomrule
\end{tabular}
}\end{adjustbox}
\caption{\textbf{Effect of extending the vanilla actor horizon from 10 to 15 steps and rule-based accessibility-tree pruning}, across all four domains of WebArena under Gemini~3~Flash. Per-domain success rate (\%) and total token usage per task (K). Values are means over three runs (mean~$\pm$~{\scriptsize std}). The last row corresponds to \budgetactor{}.}
\label{tab:pruning-ablation}
\end{table*}

The two changes play distinct roles: the extended horizon drives the success-rate gain, while pruning acts primarily as a token-reduction mechanism.
The horizon effect can be isolated: even the unpruned 15-step actor (the \emph{15 steps} row) already outperforms all three augmented methods, so \budgetactor{}'s accuracy advantage comes from the longer horizon rather than from pruning.
Pruning instead helps pay for that horizon: the tokens it saves largely offset the cost of the five additional interaction steps, so \budgetactor{} often uses no more tokens than the plain 10-step actor despite its longer horizon.
Pruning's effect on success rate is small and domain-dependent, as expected, since it removes redundant accessibility-tree text rather than adding reasoning capability.
When it does help, we suspect this is because removing duplicate entries makes the context more information-dense and shortens the token distance between related elements, such as a label and its field, which may make actionable content easier to track; the mixed results across domains indicate that this benefit is not uniform.
We also apply the same pruning procedure to the augmented baselines, with results reported in Appendix~\ref{app:baseline-pruning}.

\section{Practical advantages of stateless evaluation}
\label{sec:practical}

Beyond success rate and token cost, two practical properties differ between \budgetactor{} and the augmented methods, both stemming from the absence of cross-task state.

\spara{Parallelism.} \budgetactor{} treats tasks as independent episodes and maintains no shared cross-task state.
As a result, with $N$ workers, tasks can be executed concurrently, reducing end-to-end wall-clock time by up to a factor of $N$.
In contrast, AWM, ASI, and ReasoningBank are inherently sequential: each task must finish before the next begins because the memory or skill state derived from its trajectory is reused in subsequent tasks.
This makes parallel deployability an additional practical advantage of \budgetactor{}, beyond its token efficiency.

\spara{Fault tolerance.} Web benchmarks are run against live or simulated website environments that are not always perfectly reliable: a server may become temporarily unresponsive, a page may time out, or a browser session may crash mid-task for reasons unrelated to the agent~\citep{kara-etal-2025-waber}.
For \budgetactor{}, any such failure is local: the affected task can simply be re-run in isolation and the result substituted, since no other task depends on its outcome.
For AWM, ASI, and ReasoningBank, this repair is not possible in general.
If task $k$ fails due to an environment glitch, the knowledge state used by all subsequent tasks $k{+}1, k{+}2, \ldots$ was already formed without a potential successful trajectory from task $k$.
Re-running task $k$ in isolation does not retroactively correct the downstream state; a proper repair would require re-running the entire experiment, which can be expensive.
A single transient environment failure can therefore silently degrade the reported performance of an online augmentation method in a way that is hard to detect.
In our experiments, when we identified a failure caused by the evaluation infrastructure rather than the agent's decisions, such as a network issue or browser crash, we re-ran the affected task immediately so that the augmentation methods could still potentially induce knowledge from it, and so that the reported numbers reflect each method's performance under its intended conditions.

\section{Conclusion}

We re-evaluated online web agent augmentation under a fixed total inference budget, asking whether test-time tokens are better spent on memory, workflow, or skill modules, or directly on additional actor steps. Across four WebArena domains and three models (Gemini~3~Flash, GPT-5.4-mini, and Qwen~3.6-27B), a token-matched vanilla actor matches or surpasses AWM, ASI, and ReasoningBank in aggregate success rate while often using fewer total tokens. We observed the same trend on WorkArena-L1 with Qwen~3.6-27B, indicating that the effect extends beyond consumer web tasks to enterprise knowledge-work settings. These results do not imply that skills or workflow memory are useless: they can provide gains in domains where task structure supports reuse. Rather, our findings show that such gains are conditional and often disappear once augmentation overhead is compared against a budget-matched actor. We also show that run-to-run variance materially affects outcomes, motivating the inclusion of multi-run experiments as a core evaluation criterion rather than an optional diagnostic. Overall, our results argue for stronger evaluation hygiene in online web agent research: total test-time cost should be reported transparently, augmented agents should be compared against budget-aware vanilla baselines, and robustness analyses should accompany aggregate success rates.

\section{Limitations}

This paper reports a focused empirical comparison on WebArena under three models (Gemini~3~Flash, GPT-5.4-mini, and Qwen~3.6-27B), plus WorkArena-L1 under Qwen~3.6-27B, and several limitations follow from this scope.

Our conclusions are restricted to the benchmarks we tested and should not be generalized to all web environments.
Also, our WorkArena-L1 results are restricted to Qwen~3.6-27B and three seeds per task type.
Our benchmarks also do not place heavy demands on long-term memory: tasks in WebArena and WorkArena-L1 are largely self-contained, so information seldom needs to be carried across many episodes.
Explicitly memory-intensive environments, where solving later tasks depends on information accumulated over many earlier ones, could shift the balance toward memory modules, and evaluating the budget-matching question there is an important direction for future work.
More generally, settings with more homogeneous task distributions, longer deployment horizons, or environments built around stable high-level APIs may yield a more favorable tradeoff for augmentation methods.

Besides, our budget-matching argument applies to online augmentation only; offline methods such as SkillWeaver~\citep{zheng-etal-2025-skillweaver} and WALT~\citep{prabhu-etal-2025-walt} amortize skill-discovery cost over many uses, and a fair comparison there requires different accounting.

\section*{Acknowledgments}

This research was supported by Mitacs through the Mitacs Accelerate program and by the Fonds de recherche du Québec -- Nature et technologies (FRQNT).
This work was carried out during an internship at ServiceNow AI Research.

\bibliography{main}
\bibliographystyle{template/servicenow}

\clearpage
\appendix

\section{Implementation details}
\label{app:implementation}

\budgetactor{}, AWM, and ASI are all run using the unified codebase and prompts released by \citet{wang-etal-2025-asi} to ensure that differences in results cannot be attributed to inconsistencies in agent scaffolding, prompt formatting, or environment interaction.
The codebase uses BrowserGym~\citep{chezelles-etal-2025-browsergym} for interacting with the environment.
AWM and ASI come from the same research group~\citep{wang-etal-2024-agent, wang-etal-2025-asi}, and AWM is included as a baseline in the more recent ASI codebase.
ReasoningBank is run using the codebase released by \citet{ouyang-etal-2025-reasoningbank} due to its distinct prompt structure and agent logic.

We set the actor LLM temperature of \budgetactor{} to $0$, the same as the default value in AWM and ASI.
ReasoningBank sets the temperature to $0.7$ and we kept the default intact.
The auxiliary modules of AWM, ASI, and ReasoningBank use their respective defaults, which we do not modify.
ReasoningBank's codebase also ships a vendored WebArena with a custom evaluation pipeline that replaces the standard BrowserGym harness.
All methods in this paper are evaluated using the unmodified BrowserGym evaluation harness to ensure identical assessment criteria across configurations.

We host Qwen~3.6-27B on four NVIDIA H100 GPUs using vLLM~\citep{kwon-etal-2023-vllm}. 
We used AI assistants during implementation and manuscript preparation for code-related tasks, including generating plotting scripts and LaTeX snippets.
We also used them to assist with writing, and for polishing and shortening paragraphs.

\section{Accessibility-tree pruning}
\label{app:pruning}

The accessibility-tree simplification used in \budgetactor{} is a deterministic rule-based function with no LLM call, model fine-tuning, or learned component, and therefore no additional token cost.
The function operates on the BrowserGym-style accessibility-tree, in which each node has a role (\textit{e.g.}, \texttt{button}, \texttt{StaticText}, \texttt{heading}) and an accessible name, and is rendered as one indented line per node.

\paragraph{Pruning rule.}
A \texttt{StaticText} child node is removed when its text content is already contained verbatim in the accessible name of its immediate parent.
This handles a common pattern in web accessibility-trees, where a parent node (such as a labelled button or a list item) carries the same text as its \texttt{StaticText} child.
Removing the duplicate cuts context length without dropping information.
\texttt{StaticText} nodes whose stripped content is empty are also removed.
Non-\texttt{StaticText} nodes, node IDs (BIDs), roles, and tree structure are not modified.
The rule has one exception: a \texttt{StaticText} child is kept whenever its parent's role is one of \{\texttt{article}, \texttt{paragraph}, \texttt{heading}, \texttt{strong}, \texttt{emphasis}, \texttt{mark}, \texttt{sectionheader}\}, even when the child's text is already in the parent's name.
This avoids hollowing out actual prose blocks where the \texttt{StaticText} child is the canonical carrier of the rendered text.

\paragraph{Icon characters.}
Before comparison, Unicode private-use characters in the range \texttt{U+E600--U+E6FF} (typically icon-font glyphs that render as visually empty placeholders) are stripped from both the parent name and the \texttt{StaticText} child.
Literal \texttt{\textbackslash uXXXX} escapes that appear in the serialized tree are also stripped.
This prevents icon characters from blocking an otherwise valid redundancy match.

\section{Effect of pruning on the augmented baselines}
\label{app:baseline-pruning}

\budgetactor{} applies rule-based accessibility-tree pruning, while AWM, ASI, and ReasoningBank are run in the released configurations of their original papers without pruning.
To characterize how the baselines behave under the same preprocessing, we apply \budgetactor{}'s pruning procedure to all three of them on the Shopping domain of WebArena, under Gemini~3~Flash and Qwen~3.6-27B.
Table~\ref{tab:baseline-pruning} reports the result.

\begin{table*}[ht]
\centering
\small
\begin{tabular}{lcccc}
\toprule
 & \multicolumn{2}{c}{\textbf{Gemini~3~Flash}} & \multicolumn{2}{c}{\textbf{Qwen~3.6-27B}} \\
\cmidrule(lr){2-3}\cmidrule(lr){4-5}
\textbf{Method} & \textbf{SR (\%) $\uparrow$} & \textbf{Tok (K) $\downarrow$} & \textbf{SR (\%) $\uparrow$} & \textbf{Tok (K) $\downarrow$} \\
\midrule
AWM              & 41.18{\scriptsize $\pm$2.1} & 70.8{\scriptsize $\pm$3.0} & 41.35{\scriptsize $\pm$1.9} & 71.8{\scriptsize $\pm$4.5} \\
\quad + pruning  & 43.32{\scriptsize $\pm$0.5} & 66.9{\scriptsize $\pm$4.3} & 39.04{\scriptsize $\pm$2.3} & 68.9{\scriptsize $\pm$2.7} \\
ASI              & 44.74{\scriptsize $\pm$1.2} & 82.8{\scriptsize $\pm$0.3} & 43.67{\scriptsize $\pm$3.5} & 78.8{\scriptsize $\pm$5.5} \\
\quad + pruning  & 45.63{\scriptsize $\pm$0.8} & 71.6{\scriptsize $\pm$5.6} & 45.81{\scriptsize $\pm$2.4} & 76.7{\scriptsize $\pm$6.3} \\
RBank            & 45.45{\scriptsize $\pm$1.4} & 54.6{\scriptsize $\pm$1.3} & 39.75{\scriptsize $\pm$0.8} & 65.4{\scriptsize $\pm$0.2} \\
\quad + pruning  & 42.25{\scriptsize $\pm$4.2} & 51.1{\scriptsize $\pm$0.2} & 39.21{\scriptsize $\pm$2.2} & 59.3{\scriptsize $\pm$1.0} \\
\midrule
\budgetactor{}   & \textbf{47.77}{\scriptsize $\pm$1.9} & \textbf{45.7}{\scriptsize $\pm$0.8} & \textbf{45.99}{\scriptsize $\pm$1.4} & \textbf{58.8}{\scriptsize $\pm$1.5} \\
\bottomrule
\end{tabular}
\caption{\textbf{Augmented baselines with and without \budgetactor{}'s pruning procedure} on the Shopping domain of WebArena, under Gemini~3~Flash and Qwen~3.6-27B. Each baseline is shown in its released configuration and with the same rule-based accessibility-tree pruning used by \budgetactor{} (\emph{+ pruning}). Values are means over three runs (mean~$\pm$~{\scriptsize std}). Success rate (\%) and total token usage per task (K).}
\label{tab:baseline-pruning}
\end{table*}

Pruning reduces token usage for every baseline under both models, consistent with its role as a cost-reduction mechanism.
Its effect on success rate is mixed: it helps AWM and ASI under Gemini~3~Flash and ASI under Qwen~3.6-27B, while slightly lowering the success rate of the remaining configurations.
Even where pruning helps, the pruned baseline still trails \budgetactor{} on success rate at a higher token cost.
The closest case is ASI under Qwen~3.6-27B, whose pruned variant reaches 45.81\% against 45.99\% for \budgetactor{}, but at 76.7K versus 58.8K tokens per task, roughly 30\% more.
Because pruning does not uniformly improve success rate and would depart from the configurations released by the original authors, we keep those released configurations as the primary baselines in the main results and report the pruned runs here.

\section{Results at a higher step budget}
\label{app:step-budget}

Our main experiments cap the augmented methods at a 10-step actor horizon and \budgetactor{} at 15 steps.
To check whether the accuracy--cost pattern persists at a higher operating point, we shift the maximum step count: the augmented methods from 10 to 15 and \budgetactor{} from 15 to 20.
We run this comparison under Qwen~3.6-27B on the Shopping, Reddit, and Admin domains of WebArena.
Table~\ref{tab:step-budget} reports the result.

\begin{table*}[ht]
\centering
\small
\begin{tabular}{lcccccc}
\toprule
 & \multicolumn{2}{c}{\textbf{Shopping}} & \multicolumn{2}{c}{\textbf{Reddit}} & \multicolumn{2}{c}{\textbf{Admin}} \\
\cmidrule(lr){2-3}\cmidrule(lr){4-5}\cmidrule(lr){6-7}
\textbf{Method} & \textbf{SR (\%) $\uparrow$} & \textbf{Tok (K) $\downarrow$} & \textbf{SR (\%) $\uparrow$} & \textbf{Tok (K) $\downarrow$} & \textbf{SR (\%) $\uparrow$} & \textbf{Tok (K) $\downarrow$} \\
\midrule
\budgetactor{} - 20 steps & \textbf{46.52}{\scriptsize $\pm$0.9} & \textbf{60.8}{\scriptsize $\pm$4.1} & \textbf{47.17}{\scriptsize $\pm$3.8} & \textbf{91.6}{\scriptsize $\pm$3.3} & 51.65{\scriptsize $\pm$2.9} & \textbf{161.4}{\scriptsize $\pm$22.4} \\
AWM - 15 steps   & 41.89{\scriptsize $\pm$3.4} & 79.5{\scriptsize $\pm$3.8} & 44.03{\scriptsize $\pm$0.5} & 105.0{\scriptsize $\pm$1.4} & 47.43{\scriptsize $\pm$4.2} & 212.6{\scriptsize $\pm$6.9} \\
ASI - 15 steps   & 45.63{\scriptsize $\pm$2.9} & 96.4{\scriptsize $\pm$1.4} & 43.08{\scriptsize $\pm$1.1} & 127.4{\scriptsize $\pm$13.7} & \textbf{51.83}{\scriptsize $\pm$3.0} & 206.8{\scriptsize $\pm$11.3} \\
RBank - 15 steps & 42.24{\scriptsize $\pm$2.8} & 79.1{\scriptsize $\pm$1.4} & 41.83{\scriptsize $\pm$2.7} & 111.6{\scriptsize $\pm$5.4} & 50.18{\scriptsize $\pm$1.1} & 198.0{\scriptsize $\pm$24.2} \\
\bottomrule
\end{tabular}
\caption{\textbf{Success rate (\%) and total token usage per task (K) at a higher step budget}, under Qwen~3.6-27B on WebArena. The augmented methods use a 15-step actor horizon and \budgetactor{} a 20-step horizon. Values are means over three runs (mean~$\pm$~{\scriptsize std}).}
\label{tab:step-budget}
\end{table*}

The ordering is largely preserved at this second operating point.
\budgetactor{} has the highest success rate on Shopping and Reddit and uses fewer tokens than every augmented method in all three domains.
On Admin, ASI is nominally $0.18$ points above \budgetactor{} (51.83\% vs.\ 51.65\%), a gap far smaller than the run-level standard deviations, and it spends 28\% more tokens to reach it (206.8K vs.\ 161.4K).
Averaged over the three domains, \budgetactor{} leads ASI on both axes, at 48.6\% success and 106.2K tokens per task versus 47.4\% and 145.6K.
Granting the augmented methods more steps therefore does not let them overtake the vanilla actor: they improve, but so does \budgetactor{} under its matched increase, and both the ordering and the cost gap remain intact.

\section{Prompt templates}
\label{app:prompts}

\budgetactor{}, AWM, and ASI use the prompt templates released by \citet{wang-etal-2025-asi} without modification, except as noted below.
Since the templates are publicly available in the ASI code repository, we do not report them here.

\paragraph{Model-specific modification.}
The prompts are model-agnostic with one exception.
When running experiments with GPT-5.4-mini, we observed two behaviors that interfered with evaluation.
First, the model frequently produced action outputs with no accompanying reasoning or explanation.
Such explanations are important for augmentation methods: the induction modules parse them to extract reusable knowledge from trajectories.
Second, the model sometimes generated turns asking the user for clarification or additional input, which is not possible during automated benchmark evaluation.
To address both issues, the following text was appended to the action prompt for GPT-5.4-mini:

\begin{quote}
\itshape
You cannot ask for follow-up questions or hand back to the user.
Do your best with the information already available.
Output a summary of your reasoning and the important information you have found, followed by the best actions at this step (in one triple backticks block).
\end{quote}

\noindent It is worth noting that \budgetactor{} is less dependent on reasoning traces in the output: the vanilla agent acts only on the current observation, so in many cases explanatory text is overhead that could in principle be removed to reduce completion tokens.
This represents an additional efficiency advantage of \budgetactor{} that our numbers do not capture.
However, to ensure a fair comparison across all methods, we applied the same prompt to all methods under each model, including the above addition for GPT-5.4-mini.

\section{Failure mode analysis of augmented agents}
\label{app:failure-modes}

Section~\ref{sec:fragility} discusses the structural fragility of programmatic skills in dynamic accessibility-tree interfaces at a conceptual level.
This appendix extends that discussion with quantitative characterizations of failure modes across all three augmented methods, and presents concrete examples of tasks where augmented agents fail while \budgetactor{} succeeds.
The analysis in this section covers the Shopping, Reddit, and Admin domains under Gemini~3~Flash and GPT-5.4-mini.
Unless stated otherwise, quantities are averaged over three independent runs per domain--model pair.

\subsection{ASI: Verification confounds function success with agent recovery}
\label{app:asi-failures}

ASI induces Python functions from successful trajectories and adds them to a shared action library after a verification episode.
The intended guarantee is that only working functions enter the library.
In the verification procedure, the induced function is called as the first decision. If the function changes the environment but fails on one of its underlying actions, the agent may still recover via primitive actions and solve the task. In that case, the episode can be judged a success and the function is written to disk, without actual verification.
A common instance of this failure is when the function fails because an element ID has changed after the page reloaded for the verification episode.
Functions that require arguments whose values depend on a prior action's page state are especially prone to this: the caller cannot supply a valid BID for an element that only becomes visible after an earlier click.
The induced functions are also typically short, averaging approximately 2.4 primitive actions per function under Gemini~3~Flash and 2.5 under GPT-5.4-mini, which limits the abstraction benefit.
Some functions also hardcode task-specific values that do not generalize across tasks; for example, the function \texttt{fill\_refund\_request\_form} in the shopping library hardcodes the complaint message ``\textit{It broke after just three days of use}'' regardless of the actual reason for the refund.

Table~\ref{tab:asi-verification} quantifies the verification-confound effect.
We define a \emph{first-step verification failure} as a verification run in which the induced function fails before the second agent action.
A \emph{recovered} verification is one that was still judged correct despite that first-step failure.
First-step failures are not rare.
Under GPT, the induced function fails immediately in 72.2\% of Shopping verifications; over half of those are recovered and still produce a stored library entry.
Even under Gemini Shopping, where the first-step failure rate is lower (33.3\%), two thirds of failures recover.
In both cases the result is a function library that contains entries whose correctness at verification time was not due to the function itself.
\\

\begin{table}[ht]
\centering
\resizebox{\linewidth}{!}{
\begin{tabular}{lccc}
\toprule
\textbf{Domain} & \textbf{Attempts} & \textbf{First-step fail} & \textbf{Recov.~rate} \\
\midrule
\multicolumn{4}{c}{\textbf{Gemini~3~Flash}} \\
\midrule
Shopping & 58.0 & 19.3 & 69.0\% \\
Reddit   & 30.0 &  9.0 & 59.3\% \\
Admin    & 64.3 &  6.3 & 63.2\% \\
\midrule
\multicolumn{4}{c}{\textbf{GPT-5.4-mini}} \\
\midrule
Shopping & 44.3 & 32.0 & 54.2\% \\
Reddit   & 25.0 & 12.3 & 45.9\% \\
Admin    & 51.7 & 22.7 & 47.1\% \\
\bottomrule
\end{tabular}
}
\caption{\textbf{ASI verification failures at the first high-level function call} on WebArena. Values are averages over three runs. \emph{Attempts}: total number of function induction attempts by ASI across all tasks. \emph{First-step fail}: number of attempts where the induced function failed before the second agent action. \emph{Recov.~rate}: fraction of first-step failures where the verification episode was nonetheless judged correct, causing a potentially broken function to be stored in the shared library.}
\label{tab:asi-verification}
\end{table}

\subsection{AWM: Generic templates and induction from failed trajectories}
\label{app:awm-failures}

AWM induces abstract natural-language workflows from trajectories the judge deems successful, then retrieves and injects the closest workflow into the actor's prompt at each inference step.
Many induced workflows are too generic to add navigational value.
A stored shopping workflow for order-history aggregation reduces to: ``\textit{click the account link, click the full-order-history link, send a summary}''.
This describes the high-level intent without preserving operational details that determine correctness, such as whether pagination is needed, what the stopping condition is, or how totals are computed.
Similarly, a stored price-range workflow directs the agent to sort, toggle direction, navigate to the last page, and report the range, but omits checks for sort direction availability, filtered result sets, or whether the last-page element is reachable in the current state.
Because the workflow is injected into the actor's prompt, the agent must decide at each step whether to follow it or ignore it, a burden that smaller models are especially ill-equipped to handle, as it competes with the current accessibility-tree and task instruction.

A more fundamental problem is that AWM's induction is gated only by the judge's verdict.
Table~\ref{tab:awm-induction} shows the fraction of workflow-induction events triggered by failed trajectories.
On Shopping, nearly half of all induction events, 49.5\% under Gemini and 52.3\% under GPT, originate from failed agent trajectories.
The contamination is substantially lower for Reddit (11--19\%), which may help explain AWM's stronger performance on that domain.
Because AWM's duplicate-suppression mechanism checks only whether a new workflow semantically overlaps an existing one, it does not filter out strategies that are simply wrong: once an incorrect workflow enters the library, it can be retrieved and applied to all future tasks in the same category.

\begin{table}[ht]
\centering
\resizebox{\linewidth}{!}{
\begin{tabular}{lccc}
\toprule
\textbf{Domain} & \textbf{Induced} & \textbf{From failed (\%)} & \textbf{Final WFs} \\
\midrule
\multicolumn{4}{c}{\textbf{Gemini~3~Flash}} \\
\midrule
Shopping & 105.7 & 49.5\% & 37.3 \\
Reddit   &  45.7 & 10.9\% & 16.7 \\
Admin    &  99.7 & 42.1\% & 48.3 \\
\midrule
\multicolumn{4}{c}{\textbf{GPT-5.4-mini}} \\
\midrule
Shopping &  78.3 & 52.3\% & 33.7 \\
Reddit   &  31.3 & 19.1\% & 13.7 \\
Admin    &  50.0 & 42.0\% & 24.3 \\
\bottomrule
\end{tabular}
}
\caption{\textbf{AWM workflow-induction statistics} on WebArena, averaged per run. \emph{Induced}: total induction events triggered by the judge for a run. \emph{From failed (\%)}: fraction originating from tasks the ground-truth evaluator classified as failed. \emph{Final WFs}: distinct workflows in the final library after duplicate suppression.}
\label{tab:awm-induction}
\end{table}

\subsection{ReasoningBank: Contamination and memory over-application}
\label{app:rb-failures}

ReasoningBank stores a few short lessons from every task trajectory regardless of outcome, labeling each entry as a success or failure based on the judge's verdict.
The $K$ most semantically similar entries are retrieved at inference time and injected into the actor's context.
Table~\ref{tab:rb-corpus} reports the size and contamination of the resulting memory corpus.
On Shopping tasks, more than half of all success-labeled entries, 52.9\% for Gemini and 59.5\% for GPT, come from trajectories that failed according to the ground-truth evaluator.
These entries are framed with positive guidance language regardless of their true provenance, so the retrieval mechanism cannot distinguish between lessons that worked and lessons that did not.

\begin{table}[ht]
\centering
\resizebox{\linewidth}{!}{%
\begin{tabular}{lcccc}
\toprule
\textbf{Domain} & \shortstack{\textbf{Total}\\\textbf{mem.}} & \shortstack{\textbf{Succ.-}\\\textbf{labeled}} & \shortstack{\textbf{FP in}\\\textbf{succ.~(\%)}} & \shortstack{\textbf{FN in}\\\textbf{fail~(\%)}} \\
\midrule
\multicolumn{5}{c}{\textbf{Gemini~3~Flash}} \\
\midrule
Shopping & 518 & 96 & 52.9\% & 40.7\% \\
Reddit   & 294 & 32 & 30.2\% & 25.3\% \\
Admin    & 507 & 86 & 42.6\% & 40.4\% \\
\midrule
\multicolumn{5}{c}{\textbf{GPT-5.4-mini}} \\
\midrule
Shopping & 544 & 67 & 59.5\% & 18.4\% \\
Reddit   & 307 & 20 & 60.0\% & 17.4\% \\
Admin    & 532 & 62 & 48.1\% & 25.2\% \\
\bottomrule
\end{tabular}}
\caption{\textbf{ReasoningBank memory-corpus statistics} on WebArena, averaged over three runs. Tasks processed per run: 187 (Shopping), 106 (Reddit), 182 (Admin). \emph{Succ.-labeled}: tasks labeled successful by the judge. \emph{FP in succ.}: fraction of success-labeled tasks from trajectories that failed according to the ground-truth evaluator. \emph{FN in fail}: fraction of fail-labeled tasks from trajectories that succeeded according to the ground-truth evaluator.}

\label{tab:rb-corpus}
\end{table}

A separate concern is injection pressure.
Even though only a few memories are retrieved per task, they are referenced heavily in the agent traces: across the domain--model configurations in Table~\ref{tab:rb-corpus}, we observe an average of 8--12 explicit memory-item mentions per task, with over 95\% of Gemini tasks and over 75\% of GPT tasks referencing at least one retrieved memory.
The retrieved memories therefore function as active policy hints rather than passive background.
When a memory is overly general or mildly mismatched to the current task, the agent may prioritize it over the current accessibility-tree, producing extra navigation steps, unnecessary checks, or incorrect early termination.
\budgetactor{}, which carries no retrieved memory layer, avoids this class of distraction entirely by grounding each action in the current observation.

\subsection{Discussion: Structural limits of judge-based quality control}
\label{app:judge-limits}

The contamination patterns described above for AWM and ReasoningBank, and the verification confound in ASI, share a common structural root: each method delegates quality control to an LLM judge whose input is fundamentally incomplete for the tasks it is asked to evaluate.

The judge receives the agent's textual output (its reasoning trace and final response) and, in some configurations, a snapshot of the browser state at the end of the episode.
It does not observe the sequence of intermediate page states the agent traversed, the specific elements it interacted with, or whether information the agent reports was actually encountered during navigation or confabulated in the reasoning trace.
For tasks where correctness depends on having visited the right pages and extracted the right values, a judge working from this terminal view cannot reliably reconstruct whether the work was actually done.

This is not necessarily a matter of judge capability.
A more powerful model receiving the same partial input faces the same gap: the evidence needed to verify navigation-dependent correctness is simply absent from the judge's context.
Nor is it straightforwardly fixable by expanding what is passed to the judge.
Supplying the full trace at every step
would require passing on the order of hundreds of thousands of tokens per judge call for a ten-step task on a content-heavy page, making each evaluation call more expensive than the agent trajectory it is judging, and possibly exceeding practical context limits entirely.

\begin{table*}[t!]
\centering
\small
\resizebox{\textwidth}{!}{%
\begin{tabular}{lp{3.0cm}p{5.6cm}p{5.6cm}}
\toprule
\textbf{Method} & \textbf{Intent} & \textbf{Augmented agent behavior} & \textbf{\budgetactor{} behavior} \\
\midrule
\addlinespace
AWM
  & ``Buy the highest-rated product in Ceiling Light within a budget above \$1000.''
  & Retrieved workflow for ``Buy the highest-rated product'' prescribes sorting then clicking the top result. Agent buys the wrong product because the workflow does not account for rating ties or the budget threshold.
  & Applies sorting, checks rating and price against the budget, and selects the correct product. \\
\addlinespace
ASI
  & ``Change the delivery address for my most recent order to 3 Oxford St, Cambridge, MA.''
  & Calls \texttt{navigate\_to\_contact\_page} and \texttt{fill\_contact\_form} using the element ID from the docstring example (\texttt{'1381'}). Receives \texttt{ValueError} and \texttt{TypeError}; form submission fails.
  & Navigates to Address Book, adds the new address, recognizes that the order address cannot be changed through the UI, and reports accordingly. \\
\addlinespace
RBank
  & ``What are the top-5 best-selling products in 2023?''
  & Retrieved memory on handling ranking ties directs attention to tie-breaking. Agent misidentifies the 4th and 5th items (reports \emph{Sparta Gym Tank} and \emph{Angel Light Running Short} instead of \emph{Sprite Stasis Ball} and \emph{Hawkeye Yoga Short}).
  & Reads the same report without distraction and returns the correct ranked list. \\
\bottomrule
\end{tabular}}
\caption{\textbf{Example failures of augmented agents} where \budgetactor{} succeeds. Tasks are selected from WebArena (single runs shown; all from Gemini~3~Flash).}
\label{tab:failure-examples}
\end{table*}

\subsection{Illustrative task-level failures}
\label{app:failure-examples}

Table~\ref{tab:failure-examples} presents representative task-level failures drawn from our Gemini~3~Flash runs, showing in each case what the augmented agent does and how \budgetactor{} succeeds on the same task.

\section{Per-run variance results}
\label{app:variance}

Tables~\ref{tab:variance-gemini}, \ref{tab:variance-gpt}, and~\ref{tab:variance-qwen} report success rate and token usage for each method across three independent runs under Gemini~3~Flash, GPT-5.4-mini, and Qwen~3.6-27B, respectively.
We report sample standard deviations across the three runs.

\begin{table*}[ht]
\centering
\small
\begin{tabular}{lcccccccc}
\toprule
& \multicolumn{4}{c}{\textbf{Success Rate (\%) $\uparrow$}} & \multicolumn{4}{c}{\textbf{Tok/task (K) $\downarrow$}} \\
\cmidrule(lr){2-5} \cmidrule(lr){6-9}
\textbf{Method} & \textbf{Run a} & \textbf{Run b} & \textbf{Run c} & \textbf{Mean$\pm$Std} & \textbf{Run a} & \textbf{Run b} & \textbf{Run c} & \textbf{Mean$\pm$Std} \\
\midrule
\multicolumn{9}{c}{\textbf{Shopping}} \\
\midrule
\budgetactor{}      & 47.59 & 49.73 & 45.99 & \textbf{47.77}{\scriptsize $\pm$1.9} & 46.6 & 45.5 & 45.1 & \textbf{45.7}{\scriptsize $\pm$0.8} \\
AWM            & 43.32 & 39.04 & 41.18 & 41.18{\scriptsize $\pm$2.1} & 68.0 & 70.6 & 73.9 & 70.8{\scriptsize $\pm$3.0} \\
ASI            & 43.32 & 45.45 & 45.45 & 44.74{\scriptsize $\pm$1.2} & 82.4 & 83.0 & 83.0 & 82.8{\scriptsize $\pm$0.3} \\
RBank          & 44.39 & 44.91 & 47.05 & 45.45{\scriptsize $\pm$1.4} & 56.0 & 54.1 & 53.6 & 54.6{\scriptsize $\pm$1.3} \\
\midrule
\multicolumn{9}{c}{\textbf{Reddit}} \\
\midrule
\budgetactor{}      & 48.11 & 47.17 & 47.17 & \textbf{47.48}{\scriptsize $\pm$0.5} & 71.7 & 69.5 & 73.8 & \textbf{71.7}{\scriptsize $\pm$2.2} \\
AWM            & 49.06 & 46.23 & 47.17 & \textbf{47.48}{\scriptsize $\pm$1.4} & 87.6 & 88.9 & 85.0 & 87.2{\scriptsize $\pm$2.0} \\
ASI            & 47.17 & 43.40 & 44.34 & 44.97{\scriptsize $\pm$2.0} & 98.5 & 89.3 & 96.0 & 94.6{\scriptsize $\pm$4.8} \\
RBank          & 39.62 & 38.68 & 41.50 & 39.93{\scriptsize $\pm$1.4} & 78.9 & 73.6 & 77.8 & 76.8{\scriptsize $\pm$2.8} \\
\midrule
\multicolumn{9}{c}{\textbf{Admin}} \\
\midrule
\budgetactor{}      & 53.85 & 56.59 & 56.59 & \textbf{55.68}{\scriptsize $\pm$1.6} &  95.2 &  98.6 & 103.2 &  \textbf{99.0}{\scriptsize $\pm$4.0} \\
AWM            & 47.80 & 47.80 & 46.70 & 47.43{\scriptsize $\pm$0.6} & 136.6 & 147.0 & 144.4 & 142.7{\scriptsize $\pm$5.4} \\
ASI            & 53.85 & 51.65 & 52.75 & 52.75{\scriptsize $\pm$1.1} & 132.0 & 152.3 & 133.9 & 139.4{\scriptsize $\pm$11.2} \\
RBank          & 47.80 & 51.10 & 47.80 & 48.90{\scriptsize $\pm$1.9} & 120.8 & 126.4 & 126.5 & 124.6{\scriptsize $\pm$3.3} \\
\midrule
\multicolumn{9}{c}{\textbf{GitLab}} \\
\midrule
\budgetactor{}      & 28.33 & 29.44 & 29.44 & \textbf{29.07}{\scriptsize $\pm$0.6} & 79.8 & 76.2 & 78.0 & 78.0{\scriptsize $\pm$1.8} \\
AWM            & 25.00 & 23.33 & 25.00 & 24.44{\scriptsize $\pm$1.0} & 93.1 & 91.0 & 92.6 & 92.3{\scriptsize $\pm$1.1} \\
ASI            & 22.78 & 21.11 & 25.00 & 22.96{\scriptsize $\pm$2.0} & 102.3 & 107.0 & 113.8 & 107.7{\scriptsize $\pm$5.8} \\
RBank          & 22.78 & 23.89 & 22.22 & 22.96{\scriptsize $\pm$0.9} & 71.9 & 72.4 & 73.8 & \textbf{72.7}{\scriptsize $\pm$1.0} \\
\bottomrule
\end{tabular}
\caption{\textbf{Per-run success rate (\%) and token usage per task (K)} across three independent runs under Gemini~3~Flash on WebArena.}
\label{tab:variance-gemini}
\end{table*}

\begin{table*}[ht]
\centering
\small
\begin{tabular}{lcccccccc}
\toprule
& \multicolumn{4}{c}{\textbf{Success Rate (\%) $\uparrow$}} & \multicolumn{4}{c}{\textbf{Tok/task (K) $\downarrow$}} \\
\cmidrule(lr){2-5} \cmidrule(lr){6-9}
\textbf{Method} & \textbf{Run a} & \textbf{Run b} & \textbf{Run c} & \textbf{Mean$\pm$Std} & \textbf{Run a} & \textbf{Run b} & \textbf{Run c} & \textbf{Mean$\pm$Std} \\
\midrule
\multicolumn{9}{c}{\textbf{Shopping}} \\
\midrule
\budgetactor{}      & 37.43 & 39.04 & 39.04 & \textbf{38.50}{\scriptsize $\pm$0.9} & 55.1 & 53.5 & 54.2 & \textbf{54.3}{\scriptsize $\pm$0.8} \\
AWM            & 27.81 & 31.02 & 31.02 & 29.95{\scriptsize $\pm$1.9} & 64.2 & 61.6 & 60.2 & 62.0{\scriptsize $\pm$2.0} \\
ASI            & 34.22 & 32.62 & 32.62 & 33.15{\scriptsize $\pm$0.9} & 70.6 & 83.5 & 66.2 & 73.4{\scriptsize $\pm$9.0} \\
RBank          & 21.93 & 27.81 & 29.41 & 26.38{\scriptsize $\pm$3.9} & 59.3 & 58.2 & 56.2 & 57.9{\scriptsize $\pm$1.6} \\
\midrule
\multicolumn{9}{c}{\textbf{Reddit}} \\
\midrule
\budgetactor{}      & 38.68 & 35.85 & 29.25 & \textbf{34.59}{\scriptsize $\pm$4.8} & 89.4 & 102.5 & 96.6 & 96.2{\scriptsize $\pm$6.6} \\
AWM            & 30.19 & 30.19 & 28.30 & 29.56{\scriptsize $\pm$1.1} & 77.1 & 94.4 & 69.8 & 80.4{\scriptsize $\pm$12.6} \\
ASI            & 26.42 & 30.19 & 30.19 & 28.93{\scriptsize $\pm$2.2} & 83.8 & 85.0 & 92.8 & 87.2{\scriptsize $\pm$4.9} \\
RBank          & 18.87 & 24.53 & 22.64 & 22.01{\scriptsize $\pm$2.9} & 67.8 & 72.5 & 67.3 & \textbf{69.2}{\scriptsize $\pm$2.9} \\
\midrule
\multicolumn{9}{c}{\textbf{Admin}} \\
\midrule
\budgetactor{}      & 35.16 & 35.16 & 37.36 & \textbf{35.89}{\scriptsize $\pm$1.3} & 117.4 & 133.0 & 121.9 & 124.1{\scriptsize $\pm$8.0} \\
AWM            & 28.57 & 33.52 & 34.62 & 32.24{\scriptsize $\pm$3.2} & 129.3 &  95.9 & 114.9 & \textbf{113.4}{\scriptsize $\pm$16.8} \\
ASI            & 28.57 & 35.16 & 35.16 & 32.96{\scriptsize $\pm$3.8} & 126.1 & 146.5 & 142.9 & 138.5{\scriptsize $\pm$10.9} \\
RBank          & 36.26 & 32.97 & 33.52 & 34.25{\scriptsize $\pm$1.8} & 121.0 & 125.5 & 122.8 & 123.1{\scriptsize $\pm$2.3} \\
\midrule
\multicolumn{9}{c}{\textbf{GitLab}} \\
\midrule
\budgetactor{}      & 20.00 & 24.44 & 22.22 & \textbf{22.22}{\scriptsize $\pm$2.2} & 94.4 & 85.5 & 89.6 & 89.8{\scriptsize $\pm$4.4} \\
AWM            & 16.11 & 18.33 & 17.22 & 17.22{\scriptsize $\pm$1.1} & 92.5 & 97.0 & 97.5 & 95.7{\scriptsize $\pm$2.7} \\
ASI            & 21.11 & 21.11 & 20.00 & 20.74{\scriptsize $\pm$0.6} & 94.8 & 97.9 & 94.4 & 95.7{\scriptsize $\pm$1.9} \\
RBank          & 11.11 & 15.56 & 16.67 & 14.45{\scriptsize $\pm$2.9} & 70.9 & 70.3 & 66.8 & \textbf{69.3}{\scriptsize $\pm$2.2} \\
\bottomrule
\end{tabular}
\caption{\textbf{Per-run success rate (\%) and token usage per task (K)} across three independent runs under GPT-5.4-mini on WebArena.}
\label{tab:variance-gpt}
\end{table*}

\begin{table*}[ht]
\centering
\small
\begin{tabular}{lcccccccc}
\toprule
& \multicolumn{4}{c}{\textbf{Success Rate (\%) $\uparrow$}} & \multicolumn{4}{c}{\textbf{Tok/task (K) $\downarrow$}} \\
\cmidrule(lr){2-5} \cmidrule(lr){6-9}
\textbf{Method} & \textbf{Run a} & \textbf{Run b} & \textbf{Run c} & \textbf{Mean$\pm$Std} & \textbf{Run a} & \textbf{Run b} & \textbf{Run c} & \textbf{Mean$\pm$Std} \\
\midrule
\multicolumn{9}{c}{\textbf{Shopping}} \\
\midrule
\budgetactor{}      & 47.06 & 44.39 & 46.52 & \textbf{45.99}{\scriptsize $\pm$1.4} & 57.9 & 60.5 & 57.9 & \textbf{58.8}{\scriptsize $\pm$1.5} \\
AWM            & 39.57 & 43.30 & 41.18 & 41.35{\scriptsize $\pm$1.9} & 73.6 & 66.7 & 75.1 & 71.8{\scriptsize $\pm$4.5} \\
ASI            & 47.06 & 43.85 & 40.10 & 43.67{\scriptsize $\pm$3.5} & 72.6 & 81.1 & 82.8 & 78.8{\scriptsize $\pm$5.5} \\
RBank          & 39.57 & 40.64 & 39.04 & 39.75{\scriptsize $\pm$0.8} & 65.6 & 65.2 & 65.3 & 65.4{\scriptsize $\pm$0.2} \\
\midrule
\multicolumn{9}{c}{\textbf{Reddit}} \\
\midrule
\budgetactor{}      & 47.17 & 43.40 & 42.45 & \textbf{44.34}{\scriptsize $\pm$2.5} & 94.5 & 86.5 & 88.0 & \textbf{89.7}{\scriptsize $\pm$4.3} \\
AWM            & 44.34 & 45.28 & 39.62 & 43.08{\scriptsize $\pm$3.0} & 110.3 &  87.8 & 112.8 & 103.6{\scriptsize $\pm$13.8} \\
ASI            & 43.40 & 41.51 & 44.34 & 43.08{\scriptsize $\pm$1.4} & 118.1 & 125.9 & 117.6 & 120.5{\scriptsize $\pm$4.7} \\
RBank          & 39.62 & 41.51 & 42.45 & 41.19{\scriptsize $\pm$1.4} &  96.1 &  95.9 & 105.4 &  99.1{\scriptsize $\pm$5.4} \\
\midrule
\multicolumn{9}{c}{\textbf{Admin}} \\
\midrule
\budgetactor{}      & 49.45 & 52.20 & 50.54 & \textbf{50.73}{\scriptsize $\pm$1.4} & 141.2 & 125.3 & 124.4 & \textbf{130.3}{\scriptsize $\pm$9.5} \\
AWM            & 45.60 & 43.40 & 49.45 & 46.15{\scriptsize $\pm$3.1} & 192.8 & 172.3 & 170.3 & 178.5{\scriptsize $\pm$12.5} \\
ASI            & 51.65 & 48.35 & 47.25 & 49.08{\scriptsize $\pm$2.3} & 177.7 & 183.1 & 185.9 & 182.2{\scriptsize $\pm$4.2} \\
RBank          & 46.15 & 47.25 & 49.45 & 47.62{\scriptsize $\pm$1.7} & 160.8 & 162.9 & 147.1 & 156.9{\scriptsize $\pm$8.6} \\
\midrule
\multicolumn{9}{c}{\textbf{GitLab}} \\
\midrule
\budgetactor{}      & 27.22 & 29.44 & 27.78 & \textbf{28.15}{\scriptsize $\pm$1.2} & 109.0 & 96.2 & 100.2 & 101.8{\scriptsize $\pm$6.5} \\
AWM            & 23.33 & 22.22 & 24.44 & 23.33{\scriptsize $\pm$1.1} & 102.9 & 106.6 & 97.0 & 102.2{\scriptsize $\pm$4.8} \\
ASI            & 27.22 & 25.55 & 24.44 & 25.74{\scriptsize $\pm$1.4} & 120.4 & 117.0 & 124.3 & 120.6{\scriptsize $\pm$3.6} \\
RBank          & 20.56 & 22.78 & 19.44 & 20.93{\scriptsize $\pm$1.7} & 81.2 & 84.6 & 92.3 & \textbf{86.0}{\scriptsize $\pm$5.7} \\
\bottomrule
\end{tabular}
\caption{\textbf{Per-run success rate (\%) and token usage per task (K)} across three independent runs under Qwen~3.6-27B on WebArena.}
\label{tab:variance-qwen}
\end{table*}

\end{document}